\pdfoutput=1

\documentclass[11pt]{article}

\usepackage[final]{acl}

\usepackage{times}
\usepackage{latexsym}
\usepackage{graphicx}
\usepackage{booktabs}
\usepackage{bm}
\usepackage{multirow}

\usepackage[T1]{fontenc}

\usepackage[utf8]{inputenc}

\usepackage{microtype}

\title{Exploring Effective Information Utilization in Multi-Turn Topic-Driven Conversations}

\author{Jiatong Li \\
  The University of Melbourne \\
  \texttt{jiatongl3@student.unimelb.edu.au} \\\And
  Bin He \\
  Tencent \\
  Beijing \\
  \texttt{hebinbinhe@tencent.com} \\\And
  Fei Mi \\
   Huawei Noah's Ark Lab \\
   Beijing \\
   \texttt{mifei2@huawei.com} \\}

\begin{document}
\maketitle
\begin{abstract}
Conversations are always related to certain topics. However, it is challenging to fuse dialogue history and topic information from various sources at the same time in current dialogue generation models because of the input length limit of pre-trained language models (PLMs). In order to expand the information that PLMs can utilize, we encode topic and dialogue history information using certain prompts with multiple channels of Fusion-in-Decoder (FiD) and explore the influence of three different channel settings. In this paper, our experiments focus on a specific Chinese dataset named NaturalConv, where the conversation revolves around a piece of recent news. We thoroughly compared different dialogue models and different FiD channel settings. Empirical results show that by combining our proposed \textit{whole passage} channel with \textit{additional history} channel, our methods can achieve competitive performance on NaturalConv, making it possible to encode various information from excessively long texts. 
\end{abstract}

\section{Introduction}
Based on the Transformer framework \cite{vaswani2017attention}, large-scale pre-trained language models like mBART \cite{chipman2021mbart} and T5 \cite{raffel2020exploring} have contributed to the prosperity of dialogue systems. \\ 
However, challenges still exist, especially for multi-turn topic-driven conversations. First, it is difficult for these models to infer what people are talking about and maintain consistency in multi-turn conversations without knowing background topic information. As we know, dialogue always starts with a certain topic. People talk about their daily lives, thoughts, and recent news in their conversations. In contrast, current chit-chat dialogue systems are mostly focused on modeling multi-turn conversations, which means they implicitly learn topic information from dialogue history, while knowledge augmented approaches like information retrieval \cite{guo2016deep,severyn2015learning} are seldom used in these chit-chat models. In this case, these models usually have consistency errors and are not explainable enough.\\
Second, topic information varies in length. Although information like weather condition is quite short, some news describing an event may have a length of more than 2000 tokens, making it is technically difficult for current pre-trained language models to include the entire topic information, nevertheless to say the dialogue history. There have been works trying to extend the input length like Longformer \cite{beltagy2020longformer} and HIBERT \cite{zhang2019hibert}. However, they just relieved the problem but did not solve it. Besides, if we try to shorten the topic information by extracting summaries from these topic information, details are often omitted, leading to fact errors when the conversation goes deeper.\\
In order to solve the above challenges, we gain insights from the design of Fusion-in-Decoder (FiD) framework~\cite{izacard2021leveraging,izacard2020distilling}. FiD provided an insightful way to encode extra knowledge, which is initially used in Information Retrieval. Combining query and long passages together as multiple input channels, FiD makes these channels share the same decoder to give the output, which solves the problem of extracting knowledge from a large corpus. In our research, we propose three formats of FiD channels, \textbf{\textit{whole passage} (WP)}, \textbf{\textit{split passage} (SP)}, and \textbf{\textit{additional history} (AH)}. These asymmetric channels are initialized by different prompts and are designed to fit the characteristic of the input information category.\\
In this paper, we focus on a Chinese multi-turn topic-driven dataset, NaturalConv~\cite{wang2021naturalconv}, to conduct our experiments. This dataset is typical in the defined context, which collects 6500 news articles in 6 categories as conversation topics. Meanwhile, as a multi-turn dataset, it has an average of 20.1 utterances per dialogue, which simulates the real conversation context. Experiments show that by combining \textit{whole passage} channel with \textit{additional history} channel, the model can generate the best response.\\
To summarize, our contributions mainly lie in:
\begin{itemize}
    \item We have designed a series of candidate settings of FiD channels to fit the characteristic of the input information.
    \item We explore and compare the performance of different channel settings in multi-turn topic-driven conversation context.
    \item Experiments have shown that the combination of \textit{whole passage} channel and \textit{additional history} channel can help the model generate the best response.
\end{itemize}

\begin{figure*}
  \centering
  \includegraphics[width=1.0\textwidth]{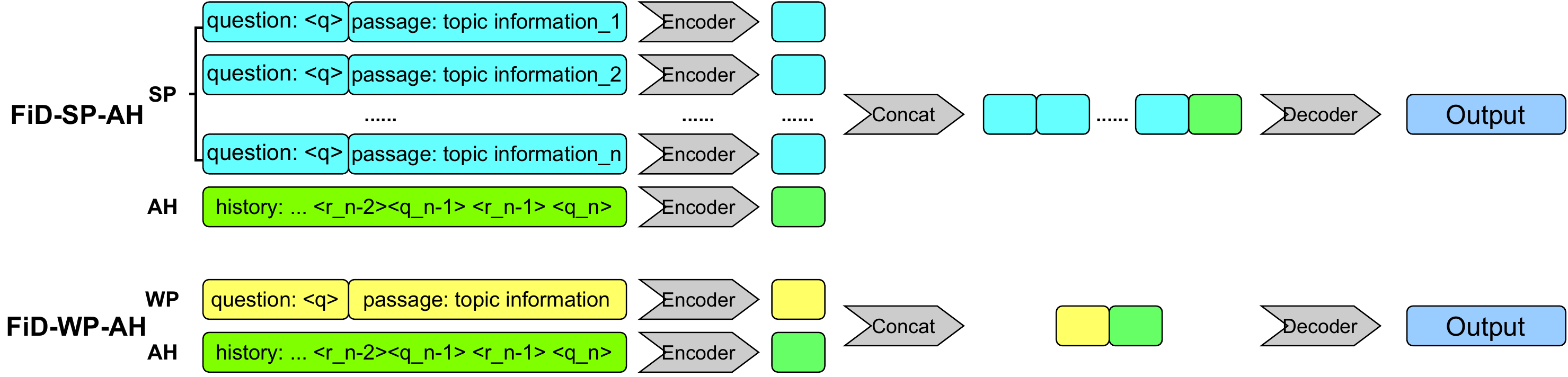}
  \caption{Overview structure of our proposed channels in FiD structure. Here, we display two different kinds of input channel combination settings, FiD-SP-AH (above) and FiD-WP-AH (below).}
  \label{fig:fig1}
\end{figure*}

\section{Related Work}
Based on GPT-2 \cite{radford2019language}, DialoGPT develops a method of modeling multi-turn conversations, which follows the characteristic of auto-regressive models~\cite{zhang2020dialogpt}. Besides, UniDS follows the schema and develops a unified dialogue system that combines chit-chat with task-oriented systems~\cite{zhao2021unids}.\\
At the same time, some models try to introduce external knowledge to dialogue systems. \citeauthor{dinan2018wizard} introduces knowledge from Wikipedia to enhance conversation generation. However, These models are grounded by static knowledge formats, which require huge human labour to collect and label structural data. Thus, Blenderbot 2.0 has proposed a new way of directly encoding information from web search~\cite{komeili2021internet,xu2021beyond}. Meanwhile, to handle knowledge from web search with GPT families, WebGPT~\cite{nakano2021webgpt} is proposed, which can generate more reliable answers with certain references.\\
For Chinese Dialogue Systems, related works are mostly focused on several datasets, including DuConv~\cite{wu2019proactive} and KdConv~\cite{zhou2020kdconv}. However, knowledge in these datasets is usually hard to organize because of the complex structures, while NaturalConv provides external knowledge as a whole passage, which best simulates the information from web search. So, we aim to make explorations based on NaturalConv and inspire future research on utilizing external information, especially web search information, in Chinese Dialogue Systems.

\section{Methods}
As shown in Figure \ref{fig:fig1}, we propose different channel formats and settings by defining certain prompts and explore valid combinations of channels to ensure that topic information and dialogue history can be encoded at the same time. In this work, we mainly define three kinds of channels, including \textbf{\textit{whole passage}}, \textbf{\textit{split passage}}, and \textbf{\textit{additional history}}. Basically, these channels are designed to fit the characteristic of the information source.
\subsection{Whole Passage}
NaturalConv provides a passage for every multi-turn conversation as the topic information. The conversation starts around the given topic news. Thus, a simple and effective method to utilize the extra topic information is to encode the topic information in a single channel, aka. \textit{whole passage}.\\
Following the format of query-and-passage pair, we use two special tokens, ``question:'' and ``passage:'', to prompt the model. Meanwhile, it should be noted that we take the current turn of conversation as the query. In this case, the model can extract related information in the passage to generate the response.
\subsection{Split Passage}
Sometimes, the length of topic information is still longer than the input length limit. Due to the length of topic information, we have to truncate the WP channel to the maximum length, which might lead to information loss. To encode external topic information at a finer granularity, we duplicate the WP channel and split the given news into several sentences. A group of WP channels will form the so-called \textit{split passage} Channel. In this case, the model can handle excessively long text. \\
However, the numbers of sentences vary from passages, while channels of FiD should be fixed for training purpose. In this case, we apply BM25 to rank the relationship between queries and sentences in the given passage.\\
Finally, top-10 sentences will be taken out of the passage and consist 10 input channels with the same query, the current turn of conversation. For passages which have less than 10 sentences, we duplicate the channel of the top sentence to fill empty channels. 
\subsection{Additional History}
In the structure of FiD, the last channel will be encoded and concatenated automatically to the tail of the embedding vector in the hidden layer, which is usually the position of the last turn conversation if we arrange the dialogue history in dialogue sequence order. \\
At the same time, auto-regressive language models generate the next sentence based on the former input. Normally, the last turn conversation might be more important than previous turns because the generation should give corresponding response to it. In this case, we assume that such property will bring latent attention patterns for pre-trained weights, which means that the tail can naturally be more important. Thus, we design a special channel, \textit{additional history} channel, to model the dialogue history.\\
Different from the above channels, this specific channel uses the token `history:' to prompt the model. Then, multi-turn dialogue history will be joined by separate tokens in dialogue sequence order. To make sure that the current query is included, this channel will be truncated at the beginning. Typically, \textit{additional history} channel will be the last channel in our combinations.
\section{Experiments}
We first make experiments on NaturalConv to test the influence of pre-trained language models, including T5-Pegasus \cite{zhuiyit5pegasus} and mBART. Meanwhile, the results will also be compared with the LSTM and Transformer model in the original paper of NaturalConv \cite{wang2021naturalconv}. \\
In this work, we select T5-Pegasus-base\footnote{https://github.com/ZhuiyiTechnology/t5-pegasus} and mBART-large, which have a 12-layer encoder and a 12-layer decoder. Meanwhile, the max input length has been set to 512. After that, we select the model that shows better performance to test the impact of different input format settings in FiD structure.\\
We use 16 Nvidia V100 (32GB) GPUs to train our models in distributed data-parallel mode and the batch size for each GPU is set to 8. Our codes and experiments are based on Pytorch structure, while the transformer structure and pre-trained weights of mBART are extracted from Hugging Face \footnote{https://huggingface.co/docs/transformers}. For more experiment settings and details, please refer to Appendix A.\\
In this paper, we calculate BLEU1-2, DIST1-2, RougeL, and F1 scores to evaluate the generation results. We mainly focus on RougeL and F1 scores, which are explained in Appendix C.\\
\begin{table*}[htb]
    \centering
    \begin{tabular}{c|c|c|c|c|c|c}
    \toprule
    Models & BLEU-1 & BLEU-2 & DIST-1 & DIST-2 & RougeL & F1 \\
    \midrule
    LSTM\cite{wang2021naturalconv}        &   \textbf{26.09} &  13.35 &   0.98 &   4.30 &    -   & 26.65   \\
    Transformer\cite{wang2021naturalconv} &   \underline{25.17} &  12.39 &   \textbf{2.91} &  15.32 &    -   & 25.73   \\
    T5-Pegasus\cite{zhuiyit5pegasus} &   22.80 &  12.57 &   2.00 &  19.64 &  27.20 & 28.58   \\
    mBART    &  25.05 &  \underline{15.94} &   1.92 &  \textbf{21.96} &  30.84 & 30.67  \\
    \multicolumn{7}{l}{\textit{mBART variants}}   \\
    Single-channel-NE    &  24.14 &  15.32 &   \underline{2.09} &  \underline{21.37} &  31.53 & 31.22  \\
    Multi-channel (FiD-SP)    &  24.20 &  15.46 &   1.85 &  18.96 &  31.05 & 30.59  \\
    Multi-channel (FiD-SP-AH) &  24.28 &  15.65 &   1.93 &  19.54 &  \underline{31.81} & \underline{31.37}  \\
    Multi-channel (FiD-WP-AH) &  24.97 &  \textbf{16.19} &   1.82 &  19.47 &  \textbf{32.38} & \textbf{31.95}  \\
    \bottomrule
    \end{tabular}
    \caption{Performance of different models and different channel settings. Here, the LSTM model consists of a two-layer encoder and a two-layer decoder, while the Transformer model has a six-layer encoder and a six-layer decoder. T5-Pegasus and mBART both have a 12-layer encoder and a 12-layer decoder. These four language models above follow the format of Single-channel-WP. The best scores are in bold, and the second-best scores are underlined. We mainly focus on RougeL and F1 scores, which are explained in Appendix C.}
    \label{table1}
\end{table*}
\subsection{Influence of pre-trained langauge models}
As shown in Table \ref{table1}, it is obvious that mBART shows the best performance among the four different language models. Here, to encode dialogue history in a single channel, the four models are all organized as Single-channel-WP, which is the concatenation of the original \textit{whole passage} channel with the \textit{additional history} channel. Each part has a maximum length limit of 256. Among the four models, mBART achieves the highest BLEU-2, F1 and, RougeL scores, which means that the generation results are the closest to the ground truth. Meanwhile, mBART also achieves a competitive DIST score, showing that the generation results have higher diversity.\\
However, it is interesting to see that although mBART achieves higher BLEU-2 and DIST-2 scores, it gets lower BLEU-1 and DIST-1 scores. In general, 2-gram scores are much harsher and represent the fluency of generated sentences, while, RougeL and F1 scores can reveal the topic relatedness to some extent. So, we believe that mBART still shows its superiority in our experiments.
\subsection{Influence of channel settings}
Table \ref{table1} also compares the results of mBART variants with different channel settings. It is clear that when combining the WP channel with the AH channel, the model generates the best responses. The full detailed illustration of how different channels are organized can be found in Appendix B.\\
To further investigate the influence from topic information, the no-evidence (NE) channel only encodes the dialogue history so that there is no topic information given to mBART as evidence.\\
Comparing the results of Single-channel-NE and mBART using Single-channel-WP, it is noticeable that mBART performs better without the topic information, which accords with the results in the original paper of NaturalConv, showing that dialogue history is more important than the topic information \cite{wang2021naturalconv}. This is reasonable because Single-channel-WP only allocates half of its maximum length to dialogue history, which leads to the loss of the former histories. The results of FiD-SP and FiD-SP-AH also prove that. Without dialogue history information, the RougeL and F1 scores of FiD-SP all have a significant drop of 0.8. \\
However, we can not tell that topic information is useless in the generation. It should be noticed that Single-channel-WP is set to only allow 256 tokens for the dialogue history, while in NaturalConv, the average length of dialogue history is close to 250 and the maximum length is more than 500, which means a large part of dialogue history has been truncated, leading to loss of information. \\
The results of FiD-WP-AH relieve the problem. By applying FiD structure, the actual maximum input length in total has been expanded to 1024 in this case, ensuring that topic information and dialogue history information could be fully utilized at the same time. The improvement is also obvious. We can see a rise of 1.44 RougeL value and 1.28 F1 value in FiD-WP-AH than Single-channel-WP.\\
It is also interesting to compare FiD-SP-AH with FiD-WP-AH. Due to the training purpose, we only select 10 sentences by BM25 for SP channels and that might miss some important information. However, it is totally acceptable because the topic information in NaturalConv only has one passage. If there is more background information from multi-source, FiD with multiple passage channels may be a better solution.

\section{Conclusion}
In this paper, we focus on NatrualConv, a Chinese multi-turn topic-dirven conversation generation corpus, to explore how to effectively utilize information with different channel settings of FiD in Multi-turn Topic Driven Conversations. By organizing channels of Fusion-in-Decoder in our proposed way, it expands the information that pre-trained language models can take, while fitting the inner characteristic of the input information, which contributes to the improvement of generation results. Our work achieves competitive results on NaturalConv and proves that by encoding topic information and dialogue history in our \textit{whole passage} and \textit{additional history} channel respectively, the model can generate the best responses.  

\section*{Acknowledgements}
We thank anonymous reviewers for their helpful opinions.

\bibliography{custom}

\appendix
\section{Hyper Parameters}
Table \ref{table3} shows the hyper parameters in our work.
\begin{table}[htb]
    \centering
    \begin{tabular}{c|c}
    \toprule
    Items & Range \\
    \midrule
    nGPUs     &   16 \\
    batch size &   8   \\
    steps     &   20000    \\
    warmup steps & 1000 \\
    lr        &   5e-4  \\
    max input length     &  512   \\
    output length & 128 \\
    \bottomrule
    \end{tabular}
    \caption{Hyper parameters}
    \label{table3}
\end{table}

\section{Full Diagram of Channels Settings}
\begin{figure*}[htb]
  \includegraphics[width=\linewidth]{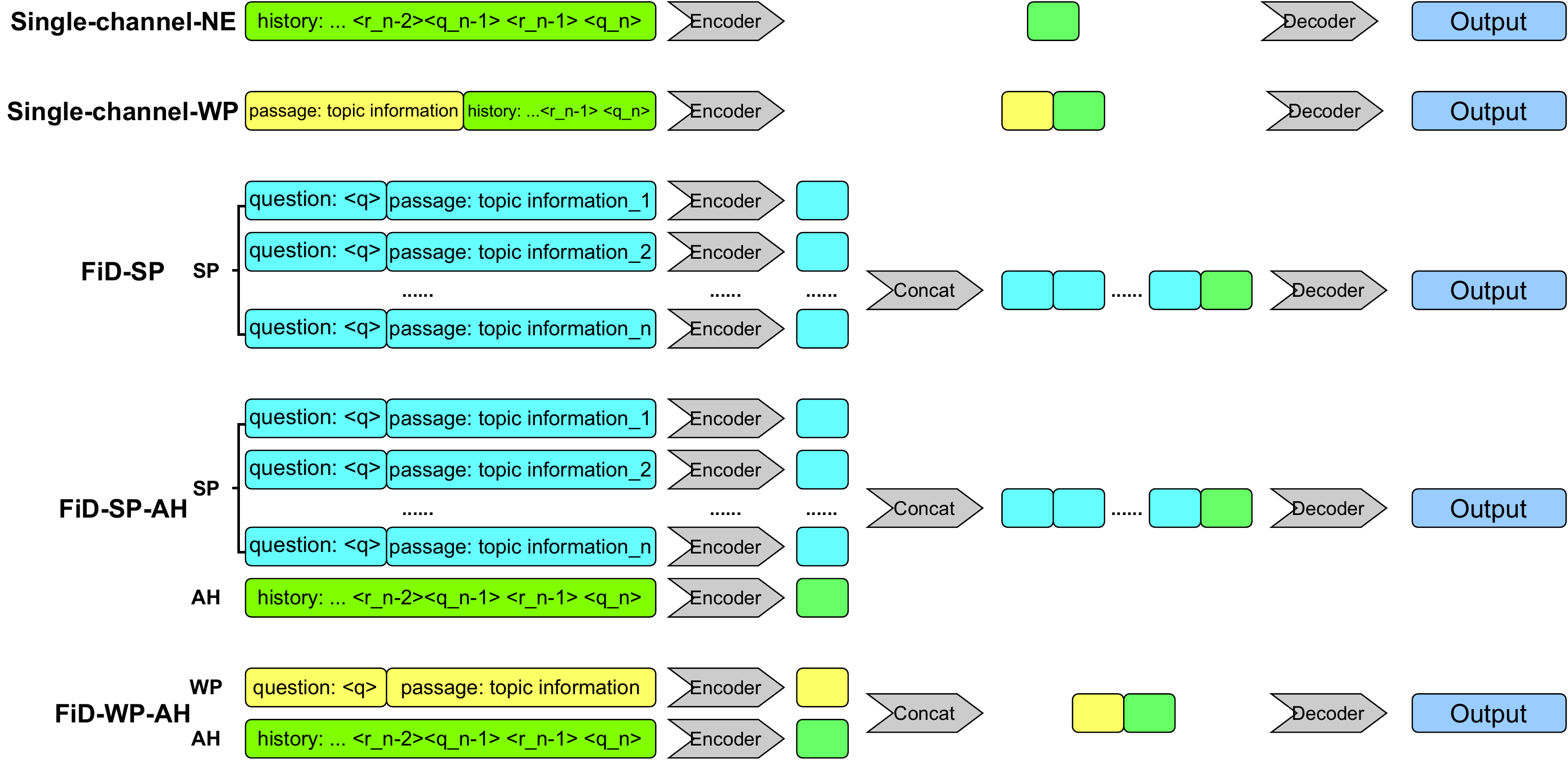}
  \caption{A full diagram to illustrate the different channel settings mentioned in this work.}
  \label{fig:fig2}
\end{figure*}

Figure \ref{fig:fig2} shows the details of different channel settings that are designed and compared in our work.

\section{Metrics}
In this work, we select BLEU1-2, DIST1-2, RougeL, and F1 scores to evaluate the generations. It should be noted that RougeL and F1 scores are more important than the others from our perspective of view. Reasons are to follow. \\
BLEU scores are usually applied in Machine Translation to show the n-gram overlap between the target and the prediction, while DIST scores are describing the diversity of the generations. However, the two metrics omit the fluency of the generated response, which is crucial in Dialogue Generation Tasks. In this case, we also calculate the RougeL for evaluation. RougeL describes the similarity based on the longest common sub-sequence, which is closely related to the fluency of generations. Besides, F1 score is also considered because it is the most obvious way to describe the similarity between the target and the generation. Overall, we think that RougeL and F1 scores can better describe the performance of our methods.

\end{document}